\pdfoutput=1

\documentclass[11pt]{article}

\usepackage{ACL2023}

\usepackage{times}
\usepackage{latexsym}
\usepackage{CJKutf8}

\usepackage[T1]{fontenc}

\usepackage[utf8]{inputenc}

\usepackage{microtype}

\usepackage{inconsolata}
\usepackage{booktabs}
\usepackage{xcolor}
\usepackage{graphicx}
\usepackage{subfigure}
\usepackage{adjustbox}
\usepackage{amsmath}
\usepackage{pifont}
\usepackage{diagbox}

\newcommand{\PreserveBackslash}[1]{\let\temp=\\#1\let\\=\temp}

\newcommand{\source}{\mathbf{x}}

\newcommand{\goldreference}{\mathbf{y}}

\newcommand{\candhypos}{\mathcal{U}}
\newcommand{\candhypo}{\mathbf{u}}

\title{Context Consistency between Training and Testing in Simultaneous Machine Translation}

\author{
    Meizhi Zhong\textsuperscript{\rm 1}\space\space
    Lemao Liu\textsuperscript{\rm }\space\space
    Kehai Chen\textsuperscript{\rm 1}\space\space
    Mingming Yang\textsuperscript{\rm }\space\space 
    Min Zhang\textsuperscript{\rm 1}\space\space \\
    \textsuperscript{\rm 1}Institute of Computing and Intelligence, Harbin Institute of Technology, Shenzhen, China \\
    {\tt 22s051052@stu.hit.edu.cn},\space\space
    {\tt lemaoliu@gmail.com},\space\space
    {\tt chenkehai@hit.edu.cn}
}

\begin{document}
\maketitle
\begin{abstract}
Simultaneous Machine Translation (SiMT) aims to yield a real-time partial translation with a monotonically growing the source-side context.
However, there is a counterintuitive phenomenon about the context usage between training and testing: 
{\em e.g.}, the wait-$k$ testing model consistently trained with wait-$k$ is much worse than that model inconsistently trained with wait-$k'$ ($k'\neq k$) in terms of translation quality.
To this end, we first investigate the underlying reasons behind this phenomenon and uncover the following two factors: 1) the limited correlation between translation quality and training (cross-entropy) loss;
2) exposure bias between training and testing. 
Based on both reasons, we then propose an effective training approach called context consistency training accordingly, which makes consistent the context usage between training and testing by optimizing translation quality and latency as bi-objectives and exposing the predictions to the model during the training. 
The experiments on three language pairs demonstrate our intuition: our system encouraging context consistency outperforms that existing systems with context inconsistency for the first time, with the help of our context consistency training approach~\footnote{Code is available at \url{https://github.com/zhongmz/ContextConsistencyBiTraining4SiMT}}. 
\end{abstract}

%
%
\section{Introduction}
Simultaneous machine translation (SiMT)~\cite{Cho2016,gu-etal-2017-learning,zhang-feng-2022-reducing,zhang-feng-2022-modeling,zhang2022wait} aims to generate a partial translation while incrementally receiving a prefix of a source sentence. A good SiMT system should not only have low {\em latency} in generation process but also yield a complete translation with {\em high quality}. SiMT has been widely used in many real-world scenarios such as multilateral organizations and international summits~\cite{ma2018stacl}. Hence, recently it has been witnessed a surge of interests in the research about SiMT~\cite{elbayad2020efficient, zhang2021universal, zhang2022wait,zhang_information-transport-based_2022}. 

\begin{figure}[t]
\centering
\includegraphics[width=2.5in]{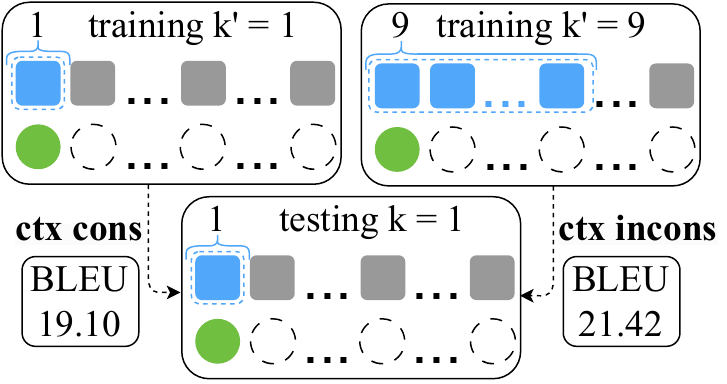}
\caption{Counterintuitive phenomenon on the context usage between training and testing: the wait-1 ($k=1$) testing model trained with $k'$=9 (denoted by ``ctx incons") outperforms the model trained with $k'$=1 (denoted by ``ctx cons") in terms of BLEU, even though the former model (trained by $k'=9$) induces a mismatch on context usage between training and testing.}
\label{fig:ills}
\end{figure}
In this paper we shed light on a {\em counterintuitive phenomenon} on the context usage between training and testing in SiMT: the wait-$k$ testing model consistently trained with wait-$k$ is worse than that model inconsistently trained with wait-$k'$ ($k'\neq k$) in terms of the evaluation metrics of SiMT, as shown in Figure~\ref{fig:ills}. 
This phenomenon was first observed by~\citet{ma2018stacl} yet without explanations. 
Subsequently, such context inconsistency training 
becomes a standard practice~\cite{elbayad2020efficient,zhang2021universal, zhang2022wait, zhang_information-transport-based_2022,guo2023glancing},  
%
even if this phenomenon is actually counterintuitive due to the mis-match between training and testing on the usage of source-side context.

To investigate the reasons behind the above counterintuitive phenomenon, we conduct experiments from two perspectives: calculating the correlation between translation quality and training (cross-entropy) loss, as well as evaluating the translation quality under the prefix-constrained decoding setting. Our empirical experiments demonstrate two reasons which takes the responsibility for the phenomenon: 1) the limited correlation between translation quality and training loss;
2) exposure bias between training and testing (\S \ref{sec.analysis}). 
Moreover, based on our findings, we then propose an effective training approach called context consistency training accordingly and break through the standard practice of inconsistent training. Its key idea is to make consistent the context usage between training and testing by optimizing translation quality and latency as bi-objectives and exposing the predictions to the model during the training. 
Our training approach is general to be applied into any SiMT systems (\S \ref{consistency-training}). 

Experiments on IWSLT14 De$\rightarrow$En, IWSLT15 Vi$\rightarrow$En and WMT15 De$\rightarrow$En utilizing several SiMT systems with two different policies, conclusively demonstrate the remarkable efficacy of our proposed approach. Our context consistency training towards bi-objectives achieves substantial gains over the original consistency training based on cross entropy. In particular, 
with the help of our training approach, our system encouraging context consistency outperforms the existing systems with context inconsistency in terms of translation quality and latency (\S \ref{experiments}). 

\paragraph{Contributions.} Our main contributions are:
\begin{itemize}
    \item We shed light on a counterintuitive phenomenon about context usage between training and testing in SiMT, and we particularly provide comprehensive explanations for this phenomenon. 
    \item Based on our explanations, we propose a simple yet effective approach, known as context consistency training, which encourages the consistent context usage between training and testing in SiMT.
    \item Our experiments conducted on three benchmarks and several SiMT systems demonstrate that our system encouraging context consistency outperforms that the existing systems with context inconsistency for the first time.
\end{itemize}

\section{Rethinking Counterintuitive Phenomenon on Context Usage}
\label{sec.analysis}

\subsection{Counterintuitive Phenomenon}
\label{sec:problem_statement}
\begin{table}[!htbp]
  \centering
  \begin{adjustbox}{width=\columnwidth}
  \begin{tabular}{c|ccccc}
    \toprule
    \diagbox{\!Train}{\! Testing} & $k$=1 & $k$=3 & $k$=5 & $k$=7 & $k$=9 \\
    \midrule
    $k'$=1 &  \underline{19.10} & 18.06 & 17.42 & 16.94 & 16.80  \\ 
    $k'$=3 & 19.29 &  \underline{23.76} & 24.97 & 25.00 & 24.40  \\ 
    $k'$=5 & 20.33 & {\bf24.89}  & \underline{26.36} & 26.93 & 27.27  \\ 
    $k'$=7 & 20.48 & 24.60 & 26.46 & \underline{27.26} & 27.81  \\ 
    $k'$=9 & {\bf21.42} & 24.82 & {\bf 26.92} & {\bf 27.84} & {\bf \underline{28.63}}  \\ 
    \bottomrule
  \end{tabular}
  \end{adjustbox}
  \caption{Evaluation by BLEU on valid set of the WMT15 De-En task for wait-$k$ policy
  . Bold: best in a column; Underline: training context is consistent to testing context.
  }
  \label{tab:problem_in_valid}
\end{table}
\paragraph{Counterintuitive Phenomenon on Valid Set}
In wait-$k$ systems, the {\bf counterintuitive phenomenon about the context usage between training and testing} was firstly observed by~\citet{ma2018stacl} yet without explanations: 
{\em the wait-$k$ testing model trained consistently with the same wait-$k$ setting is worse than the model trained with the wait-$k'$ setting ($k'\neq k$) in terms of translation quality}. 
As illustrated in Table \ref{tab:problem_in_valid}, the BLEU score obtained by the model trained with wait-$9$ surpasses the model trained with wait-$1$ by a large margin with wait-$1$ testing.
As a result, it has become a standard practice to utilize inconsistent context for training, and this practice is widely followed by~\cite{elbayad2020efficient,zhang2021universal, zhang_information-transport-based_2022, zhang2022wait,guo2022turning,guo2023glancing}, even if this phenomenon is actually counterintuitive due to the mis-match between training and testing on the usage of source-side context. 



\begin{table}[!htbp]
  \centering
  \begin{adjustbox}{width=\columnwidth}
  \begin{tabular}{c|ccccc}
    \toprule
    \diagbox{\!Train}{\! Testing} & $k$=1 & $k$=3 & $k$=5 & $k$=7 & $k$=9 \\
    \midrule
    $k'$=1 &  \underline{21.42} & 21.21 & 21.00 & 20.25 & 19.67  \\ 
    $k'$=3 & 22.07 &  \underline{25.51} & 26.73 & 26.69 & 26.33  \\ 
    $k'$=5 & 22.53 & 25.55  & \underline{27.27} & 28.06 & 28.07  \\ 
    $k'$=7 & 23.15 & 25.73 & 27.20 & \underline{28.34} & 28.63  \\ 
    $k'$=9 & {\bf23.22} & {\bf26.21} & {\bf27.52} & {\bf 28.66} & {\bf \underline{29.33}}  \\ 
    \bottomrule
  \end{tabular}
  \end{adjustbox}
  \caption{Evaluation by BLEU on training subset of the WMT15 De-En task for wait-$k$ policy.
  }
  \label{tab:problem_in_train_subset}
\end{table}

\paragraph{Counterintuitive Phenomenon on Train Subset}
One might hypothesize that this phenomenon is attributed to the generation issue from training data to valid data.
To verify this hypothesis, we conduct the similar experiments on a subset from the training data. 
We sample examples from the training data as a training subset with the same size as valid set.
Table \ref{tab:problem_in_train_subset} depicts that the situation on the training subset is almost similar to that on the valid set except for $k=3$, where the optimal $k'=9$ for the training subset rather than $k'=5$ as for the valid set. This shows that generalization from training data to valid data is not the main reason of this counterintuitive phenomenon and it is non-trivial to analyze its reasons. 
Therefore, in the next subsection, we plan to investigate the reason of this phenomenon in depth.  

\subsection{Reasons of Counterintuitive Phenomenon}

\label{sec:reason_of_problem}
\begin{table}[!htbp]
  \centering
  \begin{adjustbox}{width=\columnwidth}
  \begin{tabular}{c|cccccc}
    \toprule
    \textbf{$k$} & $1$ & $3$ & $5$ & $7$ & $9$ & $\infty$ \\
    \midrule
    \textbf{Entire} & 0.62 &	0.70 &	0.73 &	0.74 &	0.75 &	0.75 \\
    \textbf{Low} & 0.68  & 0.73 & 0.74 &  0.75 &  0.76 &  0.75 \\
    \textbf{High} & 0.27 & 0.44 & 0.51 &  0.56 &  0.60 &  0.64 \\
    \bottomrule
  \end{tabular}
  \end{adjustbox}
  \caption{Correlation between BLEU and training (cross-entropy) loss on three subsets from the training subset of the WMT15 De-En task for wait-$k$ policy, where $k=\infty$ means Full-sentence MT.
  \textbf{Entire} denotes the entire train subset, \textbf{Low} consists of those samples whose cross entropy loss is lower than the averaged loss, \textbf{High} consists of those samples whose loss is higher than the averaged loss.
  }
  \label{tab:corr_ce_loss_bleu}
\end{table}

\paragraph{Correlation between BLEU and Cross-entropy Loss in SiMT}
\label{analysis.correlation_between_BLEU_and_celoss}
Firstly, we explore the correlation between translation quality and training loss.
To investigate correlation, we measure both training loss and translation quality of each sample and calculate their Absolute Pearson Correlation in the train subset.
In the majority of SiMT systems, the training objective is based on the cross-entropy objective. 
Therefore, we assess the training loss using cross-entropy loss score in our experiments.
However, training loss is measured at the word level, while translation quality (BLEU score) is measured at the sentence level. 
To bridge this disparity, we compute the average training loss for each word within a sentence, thus representing it as sentence-level training loss. 
Table \ref{tab:corr_ce_loss_bleu} presents the results of correlation between BLEU and training (cross-entropy) loss in wait-$k$ policy. 
we reveals the following insights.
1) In wait-$k$ systems, especially when $k$ is smaller, the correlation is lower than that in Full-sentence MT. 
2) When evaluating samples with high training (cross-entropy) loss, we observe a weaker correlation (between training loss and BLEU) compared to that with low training loss. 
This observation is not difficult to understand: taking a two-class classification task as an example, if the cross-entropy loss of an example is very high (e,g., the loss is $-\log 0.2$), then the model can not predict the correct label for this example even if its loss is improved to $-\log 0.4$, because the probability of the ground-truth label is 0.4, which is less than 0.5.
{\em This suggests the reason of counterintuitive phenomenon on context usage is attributed to the relatively high cross-entropy loss for SiMT,~\footnote{Compared with full-sentence translation, SiMT uses less source-side context and thus its cross-entropy loss is higher in essense.} leading to
the weak correlation between training (cross-entropy) loss and translation quality.} 


\begin{figure}[t]
\centering
\includegraphics[width=2in]{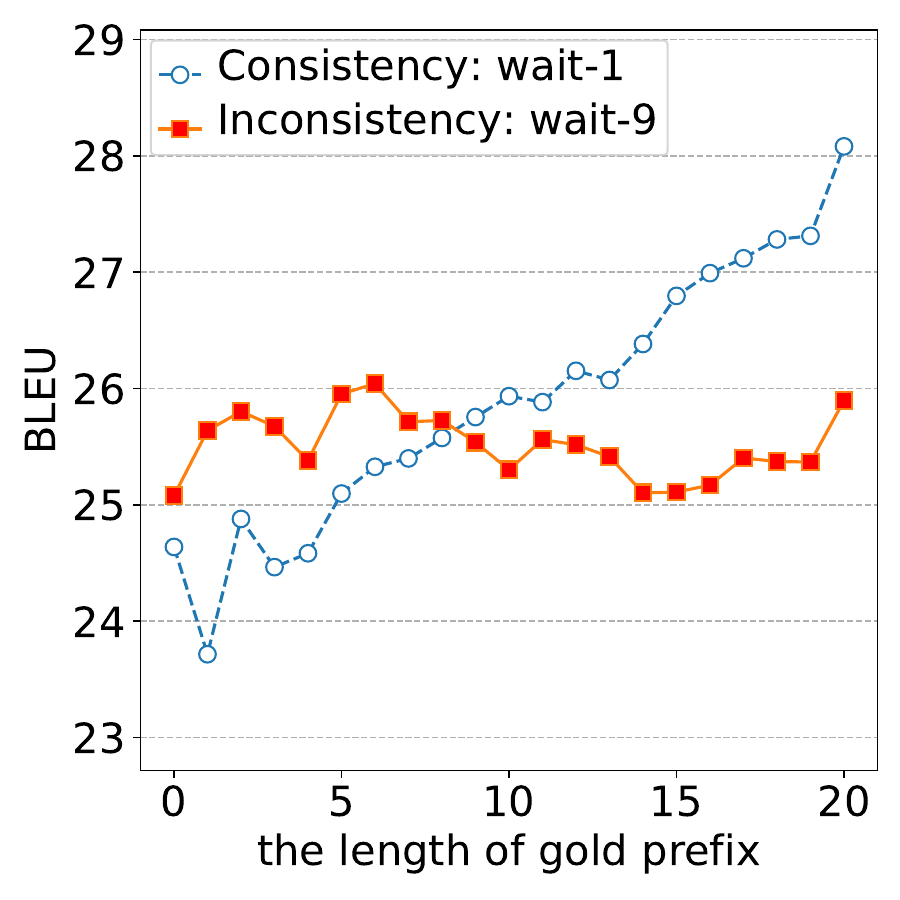}
\caption{BLEU comparison between context consistency and context inconsistency under the prefix-constrained decoding Setting. The x-aixs denotes the number of tokens for the gold prefix.}
\label{fig:prefix_wait1_vs_wait9}
\end{figure}
\paragraph{Effects of Exposure Bias on the Models Trained Consistently and Inconsistently}
\label{exposure_bias_analysis}
Since the SiMT model is typically trained by cross-entropy loss, it suffers from the well-known exposure bias, i.e., during the training the model is only exposed to the training data distribution, instead of its own predictions. Therefore, we propose to study the effects of exposure bias on the model trained with consistent context as well as the model trained with inconsistent model. 
To control the extent of exposure bias during the testing stage, we measure translation quality by BLEU for both models (e.g., the former wait-1 testing model is trained with wait-$1$ setting and the later wait-1 testing model is trained with wait-$9$ setting) under the prefix-constrained decoding setting~\cite{wuebker2016models}, where each model requires to predict the suffix for a given gold prefix. Under the this setting, as the gold prefix gets shorter, more predicted tokens are used as the context during the prefix-decoding stage and the exposure bias is more severe. 
The results as presented on Figure \ref{fig:prefix_wait1_vs_wait9} are averaged from a subset of 400 sentence pairs in the train set, all having the same number of tokens in the target (20 target tokens).
It is evident that as the gold prefix becomes shorter (i.e., exposure bias is more severe)  
the performance of the consistent model significantly deteriorates, while the inconsistent model's performance remains relatively better; however, when the number of tokens in gold prefix is larger than 10 (i.e., exposure bias is less severe), the consistent model performs better. {\em This finding reveals that one of the underlying causes of the counterintuitive phenomenon is attributed to exposure bias~\cite{ranzato2015sequence,bengio2015scheduled,zhang-etal-2019-bridging}.}


\subsection{Counterintuitive Phenomenon is Dependent on Evaluation Metrics}

\begin{table}[!htbp]
  \centering
  \begin{adjustbox}{width=\columnwidth}
  \begin{tabular}{c|ccccc}
    \toprule
    \diagbox{\!Train}{\! Testing} & $k$=1 & $k$=3 & $k$=5 & $k$=7 & $k$=9 \\
    \midrule
    $k'$=1 &  {\bf\underline{5.78}} & 5.26 & 5.00 & 4.87 & 4.81  \\ 
    $k'$=3 & 5.78 &  \underline{5.12} & 4.79 & 4.61 & 4.53  \\ 
    $k'$=5 & 5.81 & {\bf5.10}  & {\bf\underline{4.73}} & 4.53 & 4.42  \\ 
    $k'$=7 & 5.86 & 5.12 & 4.72 & \underline{4.50} & 4.38  \\ 
    $k'$=9 & 5.91 & 5.14 & 4.72 & {\bf4.49} & {\bf\underline{4.36}}  \\ 
    \bottomrule
  \end{tabular}
  \end{adjustbox}
  \caption{Evaluation by cross-entropy loss on the valid set for wait-$k$ policy.
  }
  \label{tab:problem_in_validset_loss}
\end{table}
\begin{table}[!htbp]
  \centering
  \begin{adjustbox}{width=\columnwidth}
  \begin{tabular}{c|ccccc}
    \toprule
    \diagbox{\!Train}{\! Testing} & $k$=1 & $k$=3 & $k$=5 & $k$=7 & $k$=9 \\
    \midrule
    $k'$=1 &  {\bf\underline{5.43}} & 5.11 & 4.95 & 4.87 & 4.83  \\ 
    $k'$=3 & 5.48 &  {\bf\underline{5.03}} & 4.83 & 4.73 & 4.67  \\ 
    $k'$=5 & 5.54 & 5.06  & {\bf\underline{4.81}} & 4.69 & 4.61  \\ 
    $k'$=7 & 5.60 & 5.09 & 4.82 & {\bf\underline{4.67}} & 4.59  \\ 
    $k'$=9 & 5.65 &	5.12 &	4.84 &	4.68 &	{\bf\underline{4.58}}  \\ 
    \bottomrule
  \end{tabular}
  \end{adjustbox}
  \caption{Evaluation by cross-entropy loss on the training subset for wait-$k$ policy.
  }
  \label{tab:problem_in_train_subset_loss}
\end{table}

The above both reasons motivate us to study the counterintuitive phenomenon by using the cross-entropy loss for evaluation in addition to BLEU as before, because training and testing criteria are the same and there is not exposure bias issue in this case. 
We evaluate cross-entropy loss for the wait-$k$ testing models trained with wait-$k'$ settings on the valid set and training subset.
On the valid set, we almost notice a diagonal trend, indicating the superiority of the consistent model 
, as shown in Table \ref{tab:problem_in_validset_loss}. 
On the training subset, we observe a similar diagonal trend, as illustrated in Table \ref{tab:problem_in_train_subset_loss}, indicating the counterintuitive phenomenon disappears in terms of cross-entropy loss as the evaluation metric. 
{\em These observation suggests that the counterintuitive phenomenon about context usage between training and testing is dependent on evaluation metrics, and it might be helpful to address the counterintuitive phenomenon by encouraging the consistent criterion between training and testing.}

\section{Context Consistency Training for SiMT}
\label{consistency-training}

Previous findings have shown that: 1) it is helpful to address the counterintuitive phenomenon by encouraging the consistent criterion between training and testing; 2) exposure bias is a reason for the counterintuitive phenomenon. To address the counterintuitive phenomenon and make the consistent model successful, we propose a simple and effective training approach, called context consistency training for SiMT, which not only incorporates the evaluation metrics for SiMT as training objectives (\S \ref{training_objective}) but also allows the model to expose its own predictions during training (\S \ref{generating_candidates}).
\subsection{Bi-Objectives Optimization for SiMT}
\label{training_objective}
In SiMT, the evaluation metrics of models are translation quality and latency.
Therefore, we intend to leverage both of these metrics as bi-objective in our proposed method.
\paragraph{Translation Quality} To measure translation quality of SiMT models, we employ BLEU score \cite{papineni2002bleu}.
\paragraph{Latency} Latency measurement is conducted using Average Lagging (AL) \cite{ma2018stacl}. AL quantifies the number of tokens of hypotheses that fall behind the ideal policy and is calculated as:
\begin{equation}
    \mathrm{AL}_g(\source,\candhypo) = \frac{1}{\tau} \sum_{i=1}^{\tau} g(i, \candhypo) - \frac{i-1}{|\candhypo|/|\source|}
\label{al}
\end{equation}
where 
$\tau \!=\!\mathrm{argmax}_{i}\left \{ i\mid g\left(i\right)\!=\! |\source|\right \}$, $\source$ is the source sentence, 
$\candhypo$ is the hypothesis sentence, 
and $g\left(i \right)$ is the number of waited source tokens before translating $\candhypo_{i}$ and thus it is dependent on $\candhypo_{<i}$, and its detailed definition depends on different read/write policies~\cite{ma2018stacl,zhang2022wait}. 

Formally, the SiMT model parametrized by $\theta$ can be defined as follows:
\begin{equation}
\vspace{-0.1cm}
p_g(\candhypo|\source; \theta) = \textstyle\prod_{i=1}^{|\candhypo|} p(\candhypo_i | \source_{\leq{g(i)}},\, \candhypo_{<i})
\label{eq:ourgoal}
\end{equation}
\noindent where $\candhypo$ denotes a complete translation hypothesis and $\candhypo_{<i}$ denotes its partial prefix with $i$ tokens. 

Inspired by Minimum Risk Training (MRT) \cite{shen2016minimum,wieting2019beyond}, we directly optimize the SiMT model towards its bi-objectives (i.e., BLEU and Latency) as follows: 
\begin{equation}
\mathcal{L}_g =
\sum_{\candhypo \in \candhypos(\source)} \operatorname{cost}_g(\source,\goldreference, \candhypo) \frac{p_{g}(\candhypo|\source; \theta)}{\sum_{\candhypo' \in \candhypos(\source)} p_{g}(\candhypo'|\source; \theta)}
\label{mrt_loss_function}
\end{equation}
\noindent where $\candhypos(\source)$ is a set of candidate hypotheses, $\goldreference$ is the reference and $\operatorname{cost}_g(\source,\goldreference, \candhypo)$ consists of bi-objectives: 
\begin{multline}
    \operatorname{cost}_g(\source,\goldreference, \candhypo) = \gamma\cdot\text{AL}_g(\source,\candhypo)+ \\
    (1-\gamma)\cdot(1-\text{BLEU}(\goldreference, \candhypo))
\label{bleu_al.mrt_cost}
\end{multline}
The parameter $\gamma$ is adjustable and allows us to fine-tune for different latency requirements.

\paragraph{Remark}
In~\citet{shen2016minimum,wieting2019beyond}, the cost is directly defined on a translation candidate $\candhypo$ and thus it is trivial to calculate the cost for a given $\candhypo$. However, in our scenario,  $\text{AL}_g(\source,\candhypo)$ is not only dependent on $\candhypo$ but also dependent on $g(i)$ specified by the read/write policy used in the SiMT system. As a result, during the training process, for each candidate $\candhypo$ generated via decoding, we access the SiMT model to incrementally compute the $g(i)$ for all $i$ and then compute $\text{AL}_g(\source,\candhypo)$ based on all $g(i)$ for $\candhypo$.

\subsection{Generating $n$ Candidates for Training SiMT}
\label{generating_candidates}
In the conventional training SiMT with cross-entropy loss, it does not involve the multiple candidates by decoding. 
In our scenario, to calculate the objective function defined in~\eqref{mrt_loss_function}, we have to generate a set of candidates $\candhypos$ via decoding which also allows the SiMT model to expose to the predictions and thereby alleviates exposure bias during the training stage. 
To this end, we try two different ways (Beam search and Sampling search) ~\cite{holtzman2019curious} to generate $n$-best candidates in SiMT.
Beam search is a maximization-based decoding technique that optimizes output by favoring high-probability tokens. It is widely-used in the generation of Full-sentence MT.
Sampling search~\cite{holtzman2019curious} is a stochastic decoding approach that samples from the top-$p$ portion of the probability distribution. This method excels in enhancing candidate diversity. In our experiments, we generate a set of $5$-best candidates and select 0.8 for top-$p$ in sampling search.

Furthermore, in order to calculate the $\text{AL}_g(\source,\candhypo)$ of candidates defined in Eq.~\eqref{al} which is dependent on the $g(i)$, we maintain both model score $p_g$ as well as $g(i)$ (the number of waited source words before translating $\candhypo_i$) at each timestep $i$. 
Specifically, during the decoding process, the SiMT model uses the value of $g(i)$ to incrementally specify the source context and produce the next predictive distribution $p_g$. From this predictive distribution $p_g$, we select the top $n$-best (for beam search method) or sample $n$ (for sampling method) partial candidates along with their respective $g(i)$ values.

Following~\citet{edunov2017classical,wieting2019beyond}, we employ the two-step training paradigm to train SiMT to speed up the training process: 
we first train the SiMT model with the standard cross-entropy loss, and then, in our context consistency training, we fine-tune the model by optimizing the bi-objectives (translation quality and latency) with the generated $n$-best candidates.
It is worth noting that we only generate $n$ candidates in training stage but in testing stage the greedy search is used because of the essence of SiMT.

\section{Experiments}
\label{experiments}
\subsection{Datasets}
\label{subsection.datasets}
We conduct experiments on the following datasets, which are the widely-used SiMT benchmarks.

\textbf{IWSLT14 German $\!\rightarrow \!$  English (De$\rightarrow$En)}~\cite{Cettolo14iwslt}
we train on 160K pairs, develop on 7K held out pairs and test on TED dev2010+tst2010-2013 (6,750 pairs).
Following the previous setting~\cite{elbayad2020efficient}, all data is tokenized and lower-cased and we segment sequences using byte pair encoding \cite{Sennrich16acl} with 10K merge operations. The resulting vocabularies are of 8.8K and 6.6K types in German and English respectively. 

\textbf{IWSLT15\footnote{\url{nlp.stanford.edu/projects/nmt/}} Vietnamese $\!\rightarrow \!$  English (Vi$\rightarrow$En)}~\cite{Luong15iwslt}
we train on 133K pairs, develop on TED tst2012 (1,553 pairs) and test on TED tst2013 (1,268 pairs). 
The corpus is simply tokenized by SentencePiece~\cite{kudo_sentencepiece_2018} resulting in 16K and 8K word vocabularies in English and Vietnamese respectively.

\textbf{WMT15\footnote{\url{www.statmt.org/wmt15/translation-task}} German $\!\rightarrow\! $ English (De$\rightarrow$En)}~\cite{callison-burch-etal-2009-findings} is a parallel corpus with 4.5M training pairs. We use newstest2013 (3003 pairs) as the dev set and newstest2015 (2169 pairs) as the test set. The corpus is simply tokenized by SentencePiece~\cite{kudo_sentencepiece_2018} resulting in 32k shared word vocabularies.

\subsection{System Settings}
\label{subsection.simt_system_settings}

\paragraph{SiMT with Two Policies} We conduct experiments on two kinds of SiMT systems including two different policies.
The fixed read/write system ({\bf wait-$k$ policy}) \cite{ma-etal-2019-stacl}, which first reads $k$ source words, and then alternately reads one word and writes one word. 
The adaptive read/write system ({\bf wait-info policy})~\cite{zhang2022wait} that formulates the decision of waiting or outputting is made based on the comparison results between the total information of previous target outputs and received source inputs.




The implementation of all systems are based on Transformer~\cite{vaswani2017attention} and adapted from Fairseq Library~\cite{ott-etal-2019-fairseq}. 
Following~\citet{ma2018stacl,elbayad2020efficient}, we apply Transformer-Small (4 heads) for IWSLT15 Vi$\rightarrow$En and IWSLT14 De$\rightarrow$En, Transformer-Base (8 heads) for WMT15 De$\rightarrow$En.
To avoid the recalculation of the encoder hidden states when a new source token is read, unidirectional encoder~\cite{elbayad2020efficient} is proposed to make each source token only attend to its previous words.


\paragraph{Baseline Training Approaches}


The conventional training approach of SiMT systems is the context consistency training based on cross-entropy, which is studied in \citet{ma2018stacl} and is denoted by {\bf Consistency-CE}.
In contrast, the context inconsistency training, also based on cross-entropy, involves the inconsistent context usage between training and testing stages.
This training approach is denoted by {\bf Inconsistency-CE}.
Additionally, we implement a recently widely-used special case of context inconsistency training, termed {\bf Inconsistency-CE-MP}. This method employs a multipath sampling training approach based on cross-entropy \cite{elbayad2020efficient,zhang2022wait}.

\paragraph{Our Training Approaches}
To compare our proposed systems against baselines, we follow the standard bi-objective (translation quality and lentency) evaluation paradigm for SiMT~\cite{ma2018stacl} and report BLEU \cite{papineni-etal-2002-bleu} for translation quality and Average Lagging (AL) \cite{ma-etal-2019-stacl} for latency mentioned in \S \ref{training_objective}. Our proposed context consistency training is based on bi-objectives and thereby is denoted by {\bf Consistency-Bi}, and we also implement the context consistency training based on BLEU as the uni-objective which is denoted by {\bf Consistency-Uni} for further comparison.  
For generating $n$ candidates, we implement Beam search in most cases, with the exception of the wait-$k$ policy on WMT15 De$\rightarrow$En, for which we utilize the Sampling search strategy.

\subsection{Main Results}
\begin{figure*}[t]
\centering
\subfigure[IWSLT14 De$\rightarrow$En]{
\includegraphics[width=2.0in]{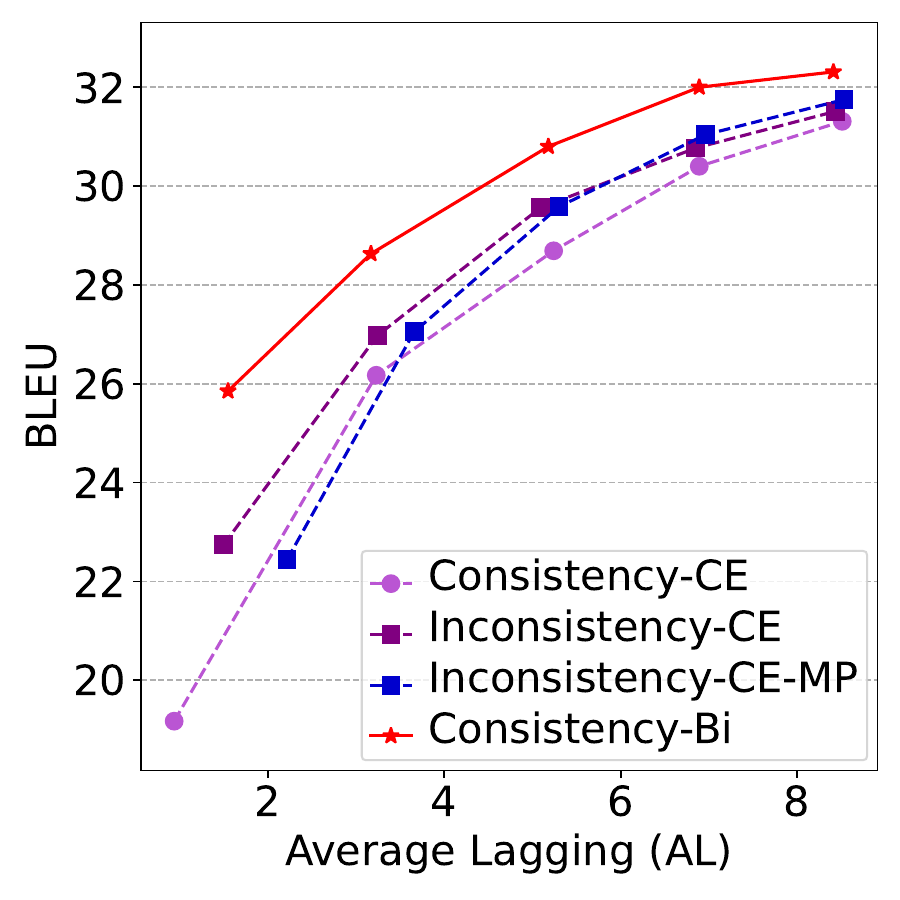}
}
\subfigure[IWSLT15 Vi$\rightarrow$En]{
\includegraphics[width=2.0in]{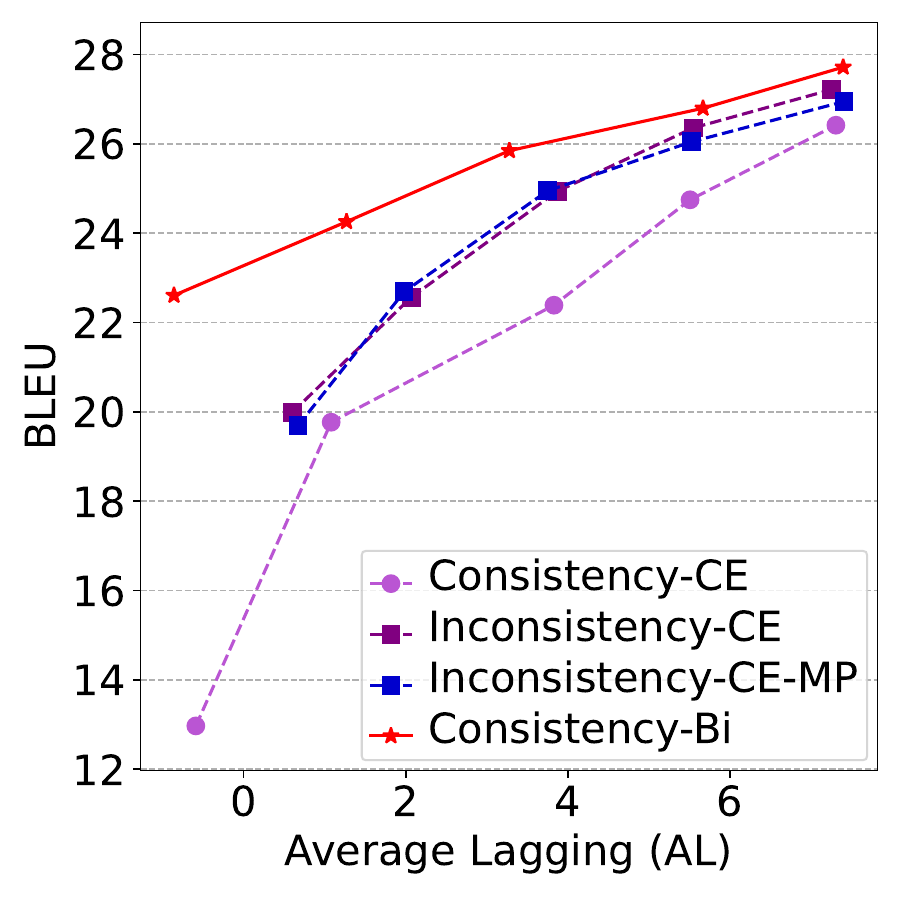}
}
\subfigure[WMT15 De$\rightarrow$En]{
\includegraphics[width=2.0in]{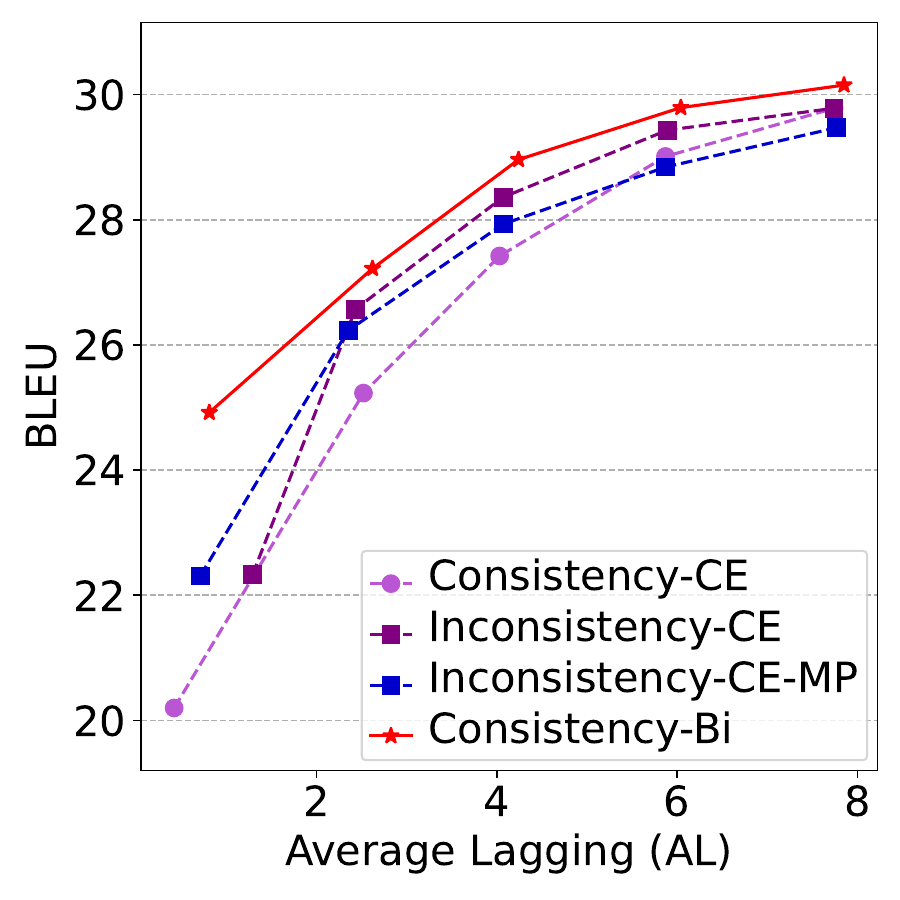}
}
\caption{Translation quality (BLEU) v.s. latency (Average Lagging, AL) in Wait-$k$ Policy.}
\label{main.waitk}
\end{figure*}


\begin{figure*}[t]
\centering
\subfigure[IWSLT14 De$\rightarrow$En]{
\includegraphics[width=2.0in]{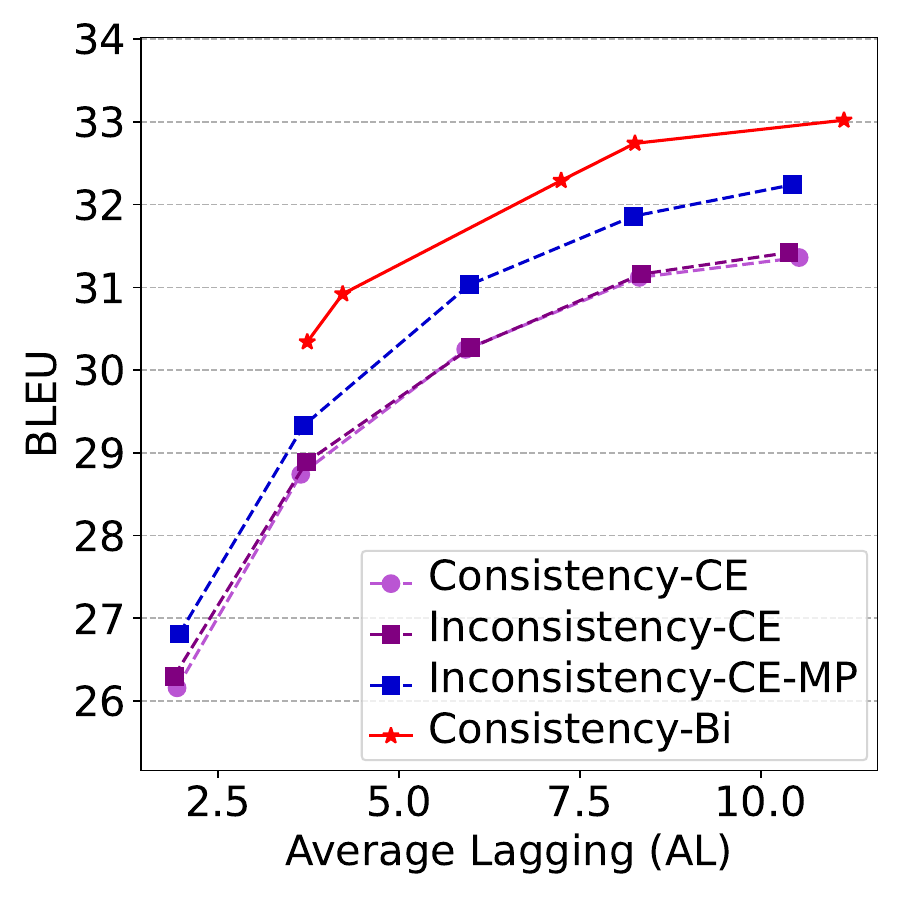}
}
\subfigure[IWSLT15 Vi$\rightarrow$En]{
\includegraphics[width=2.0in]{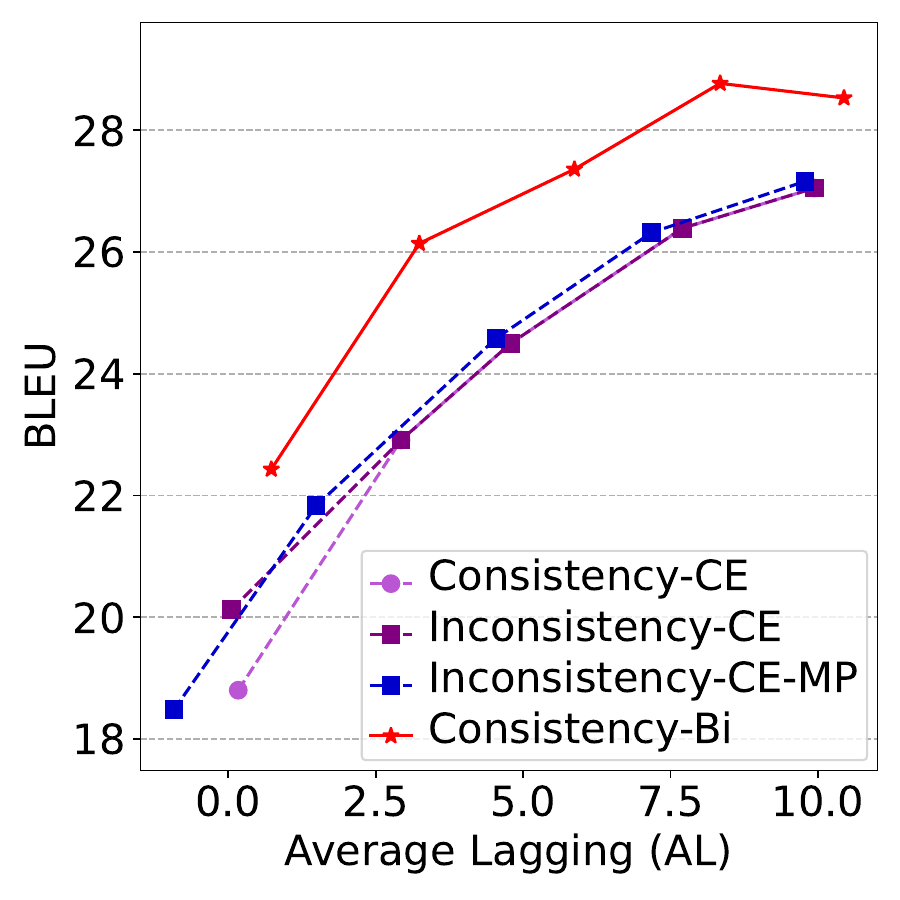}
}
\subfigure[WMT15 De$\rightarrow$En]{
\includegraphics[width=2.0in]{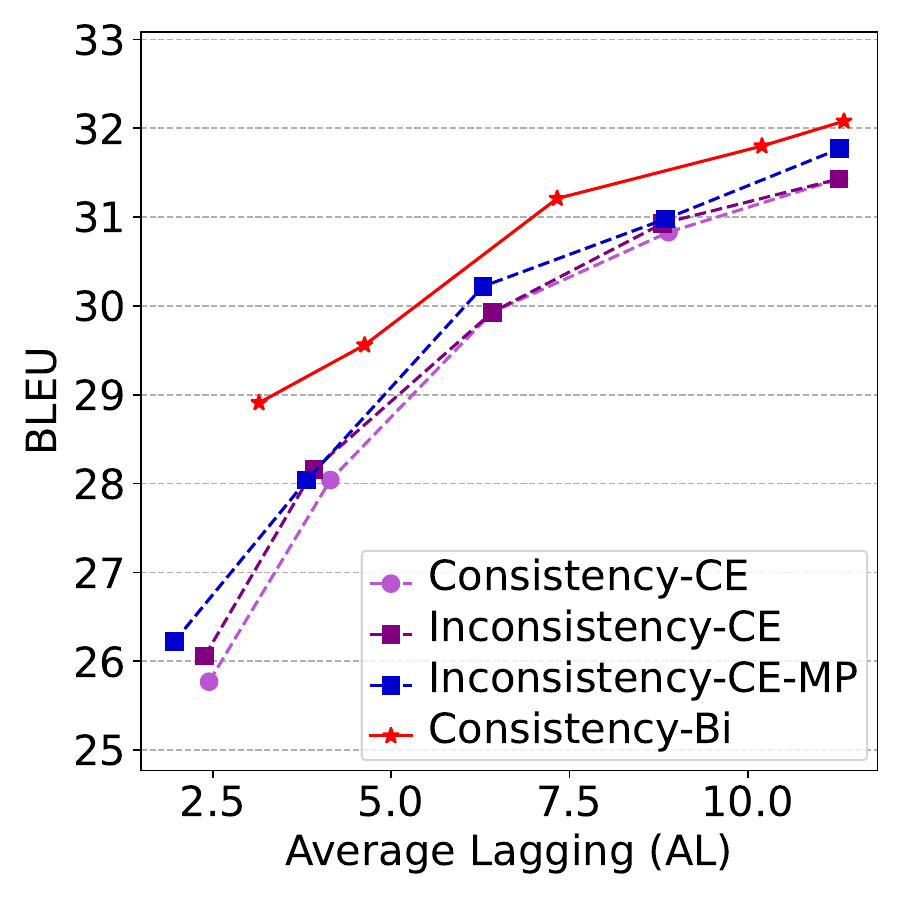}
}
\caption{Translation quality (BLEU) v.s. latency (Average Lagging, AL) in Wait-info Policy.}
\label{main.wait-info}
\end{figure*}
The results are illustrated in Figure \ref{main.waitk} and Figure \ref{main.wait-info}.
Within our proposed context consistency training approach (Consistency-Bi), all implemented SiMT systems (wait-$k$ and wait-info) exhibit significant improvements in both translation quality and latency, as evidenced by an increase in BLEU score and a decrease in AL across all the benchmarks.
This reveals that our proposed methods not only yield substantial performance improvements but also demonstrate strong generalization capabilities for SiMT systems. 

\paragraph{Wait-$k$ Policy} 
In contrast to the original consistency training (Consistency-CE), our proposed Consistency-Bi achieves over 5 BLEU improvement at low latency ($k$=1) across all datasets.
Specifically, our method improves 2.68 BLEU on IWSLT14 De-En task, 4.39 BLEU on IWSLT15 Vi-En task and 1.91 on WMT15 De-En task, respectively (average on all latency).
Furthermore, compared with inconsistency training (Inconsistency-CE and Inconsistency-CE-MP), the proposed method also demonstrates significant improvements, especially at low latency ($k$=1), achieving over 3 BLEU score increase.
This suggests that incorporating our proposed context consistency training enables a wait-$k$ model trained consistently under the same wait-$k$ testing setting is able to outperform an inconsistently trained model.

\paragraph{Wait-info Policy}
To evaluate whether our method could achieve improvements with advanced adaptive SiMT systems, we apply our proposed training method to wait-info policy \cite{zhang2022wait}.
The results are depicted in Figure~\ref{main.wait-info}. 
Similarly, in comparison to three baseline training methods, we observe significant enhancement in translation quality across all latency.
However, in IWSLT15 Vi-En task and WMT15 De-En task, Inconsistency-CE and Inconsistency-CE-MP is not significant better than Consistency-CE.
This can be attributed to the advanced read/write policy utilized by the wait-info policy, which makes more informed read/write decisions based on information.

\subsection{Ablation Study}
\begin{figure}[t]
\centering
\includegraphics[width=1.8in]{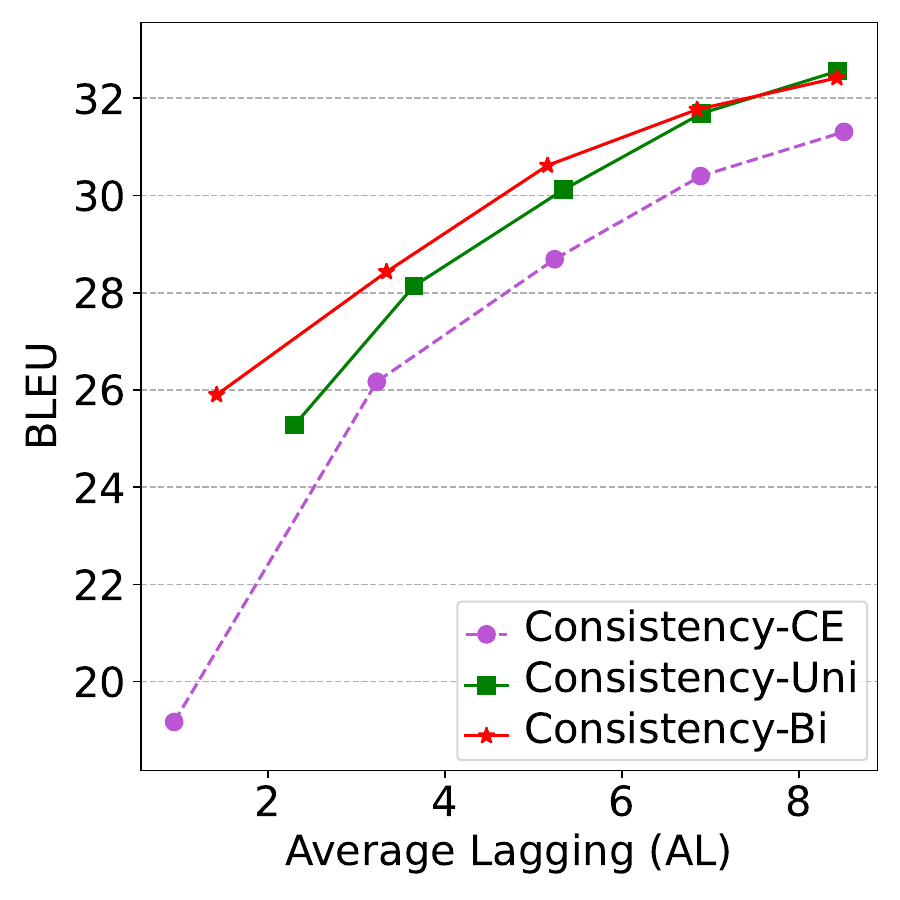}
\caption {Ablation studies between Consistency-Bi and Consistency-Uni on test set of IWSLT14 De-En task of wait-$k$ policy. }
\label{abs.uni_vs_bi}
\end{figure}
\paragraph{Consistency-Bi v.s. Consistency-Uni}
In order to validate the effectiveness of Consistency-Bi, we perform the ablation studies on Consistency-Bi (BLEU and AL) and Consistency-Uni (BLEU only) in Figure \ref{abs.uni_vs_bi}.
The experiments reveal that, compared with Consistency-Uni, Consistency-Bi not only results in lower latency but also yields superior translation quality, especially in low latency scenario ($k$=1).
This is largely attributed to the latency optimization as part of the training objective defined in \eqref{bleu_al.mrt_cost}.

\begin{figure}[h]
\centering
\subfigure[Wait-$k$ Policy]{
\includegraphics[width=1.44in]{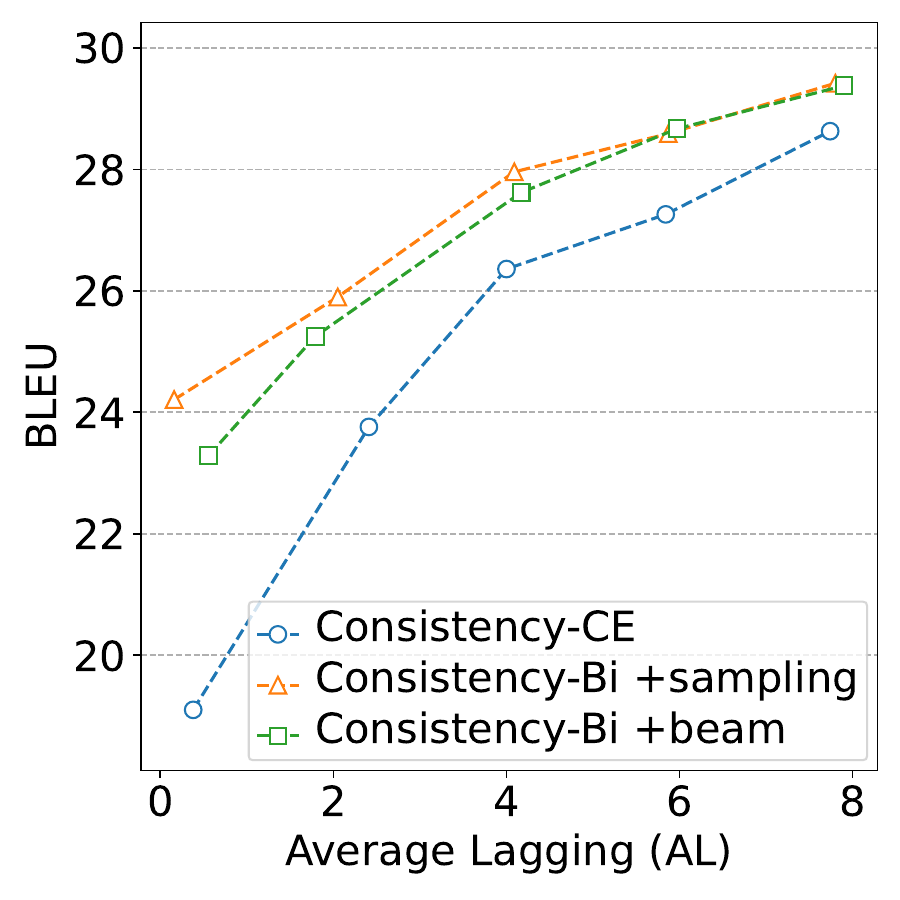} 
\label{fig:beam_vs_sampling_in_waitk}}
\hspace{-2mm}
\subfigure[Wait-info Policy]{
\includegraphics[width=1.44in]{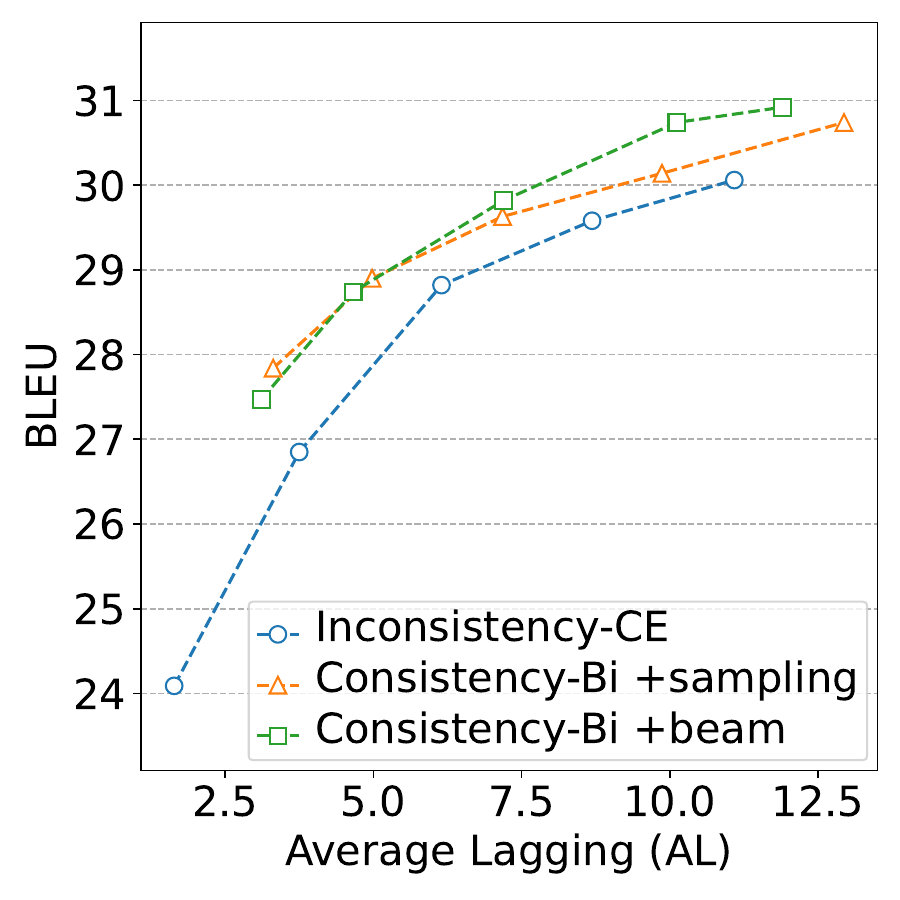} 
\label{fig:beam_vs_sampling_in_wait_info}
}
\caption{Ablation studies on $n$-best candidates generations (Beam search and Sampling search) on valid set of WMT15 De-En.}
\label{abs.beam_vs_sampling}
\end{figure}
\paragraph{Ablation studies on $n$-best candidates generations}
We conduct the ablation studies on two types of $n$-best generation methods (Beam search and Sampling search) under both wait-$k$ and wait-info policies, as depicted in Figure \ref{abs.beam_vs_sampling}.
The results reveal that under wait-$k$ policy, the performance of Consistency-Bi using sampling search are slightly superior to that using beam search. Conversely, under wait-info policy, employing beam search yields slightly better results compared to sampling search. 
Overall, the effects of beam search and sampling search on the two SiMT systems are not significantly different, suggesting that the choice of generation method is not notably sensitive within our proposed method.

\begin{table}[!htbp]
  \centering
  \begin{adjustbox}{width=\columnwidth}
  \begin{tabular}{cccccccc}
    \toprule
    $\gamma$ & $0.0$ & $0.1$ & $0.2$ & $0.3$ & $0.4$ & $0.5$ & $0.6$ \\
    \midrule
    \textbf{BLEU} & 23.5 &	23.37 &	23.08 &	23.56 &	\textbf{24.21} &	21.09 & 17.74 \\
    \textbf{AL} & 1.68  & 1.62 & 1.53 &  1.14 &  0.16 &  -1.48 & \textbf{-2.93} \\
    \bottomrule
  \end{tabular}
  \end{adjustbox}
  \caption{Ablation studies on various $\gamma$ in wait-$1$ training with wait-$1$ testing (Consistency-Bi on wait-$k$) on valid set of WMT15 De-En task.}
  \label{tab.diff_gamma}
\end{table}
\paragraph{Variation in hyperparameter $\gamma$}
Fine-tuning hyperparameter $\gamma$ defined in \eqref{bleu_al.mrt_cost} aims to achieve a better trade off between BLEU and latency in our proposed Consistency-Bi. 
As illustrated in Table \ref{tab.diff_gamma}, as $\gamma$ increases, latency (AL) decreases while the BLEU score improves, reaching its peak at $\gamma$ = 0.4.
This indicates that our proposed method can simultaneously optimize two objectives (BLEU and AL), and can achieve a value that is relatively optimal balance between BLEU and AL.

\subsection{Analysis}
\begin{figure}[t]
\centering
\subfigure[orgin(train w/ CE-obj)]{
\includegraphics[width=1.44in]{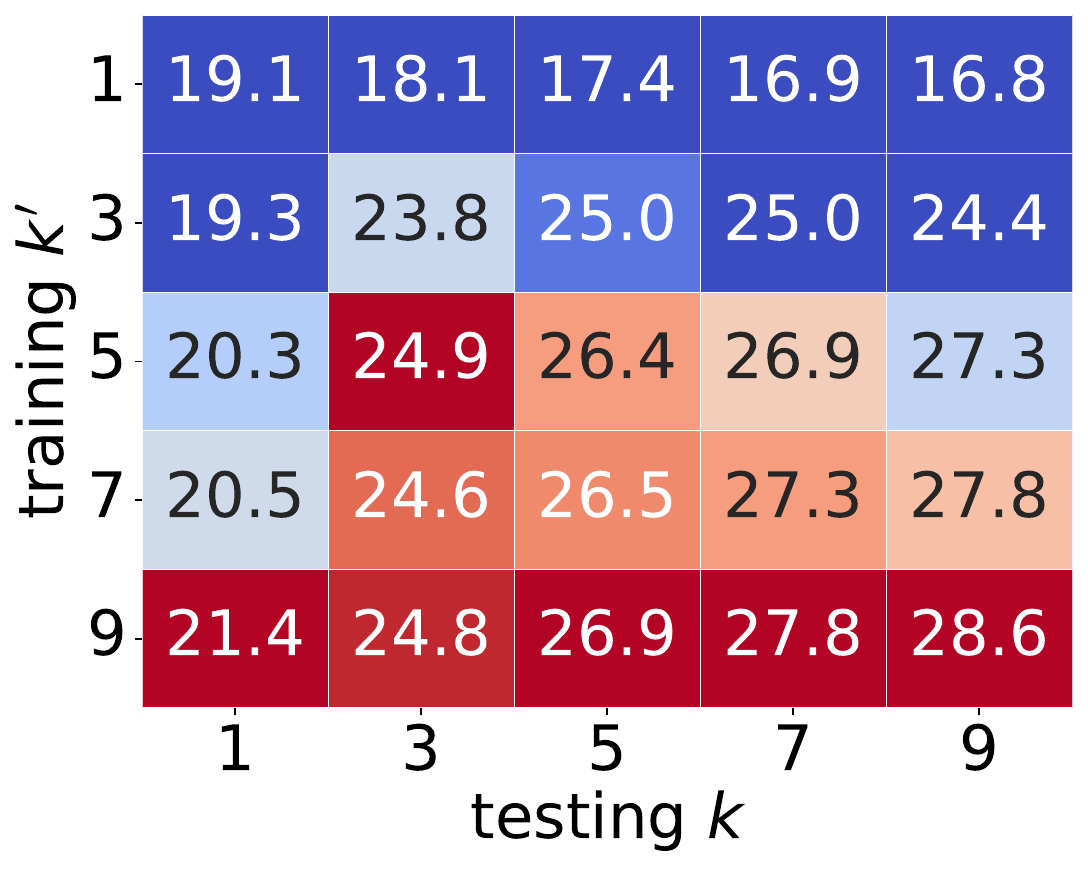} 
\label{fig:base.heatmap}}
\hspace{-2mm}
\subfigure[proposed(train w/ Bi-obj)]{
\includegraphics[width=1.44in]{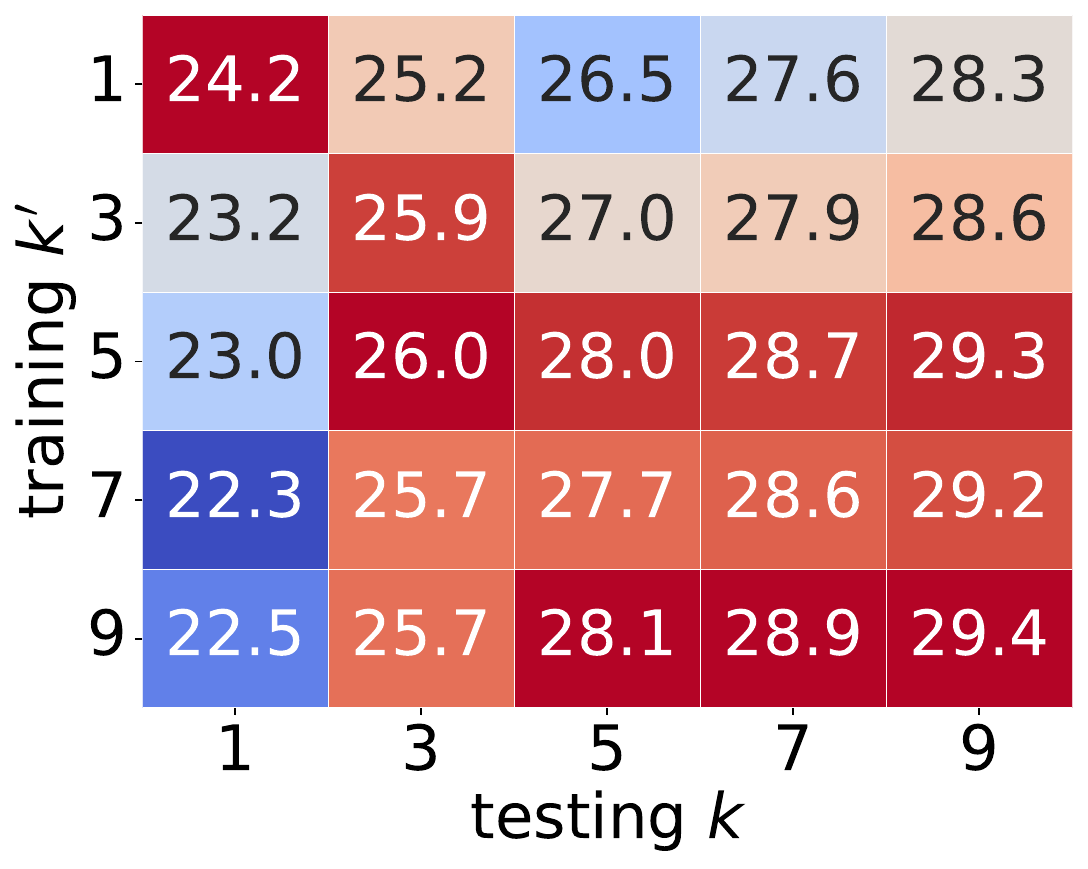} 
\label{fig:bleu-al.heatmap}
}
\caption{BLEU score comparison between the original and proposed training methods using the wait-$k'$ policy during training and wait-$k$ policy during testing on the valid set of the WMT15 De-En task. 
The diagonal line indicates consistency between training $k'$ and testing $k$.}
\label{analysis.consistency_comparison}
\end{figure}
\paragraph{Counterintuitive Phenomenon Mitigation} 
To investigate whether the counterintuitive phenomenon discribed in \S\ref{sec:problem_statement} has been alleviated, we conduct experiments using models trained with wait-$k'$ but tested with wait-$k$, as illustrated in Figure \ref{analysis.consistency_comparison}.
The results of original training method as presented in Figure \ref{fig:base.heatmap}. Optimal results for testing with $k$ are generally achieved when $k'$=9, excepted for $k$=3, where $k'$=5 yields the best.
In contrast, our proposed training method demonstrates that the best results tested with wait-$k$ closely match with the diagonal line as depicted in Figure \ref{fig:bleu-al.heatmap}.
Specifically, when testing with $k$=1 and 9, the best results match the models trained with the same value of $k'$.
For $k$=3, 5, and 7, although the best results come from different models, the differences are not significant.
These findings suggest that our method exhibits improved consistency between training and testing compared with orginal training method.

\begin{figure}[t]
\centering
\includegraphics[width=1.8in]{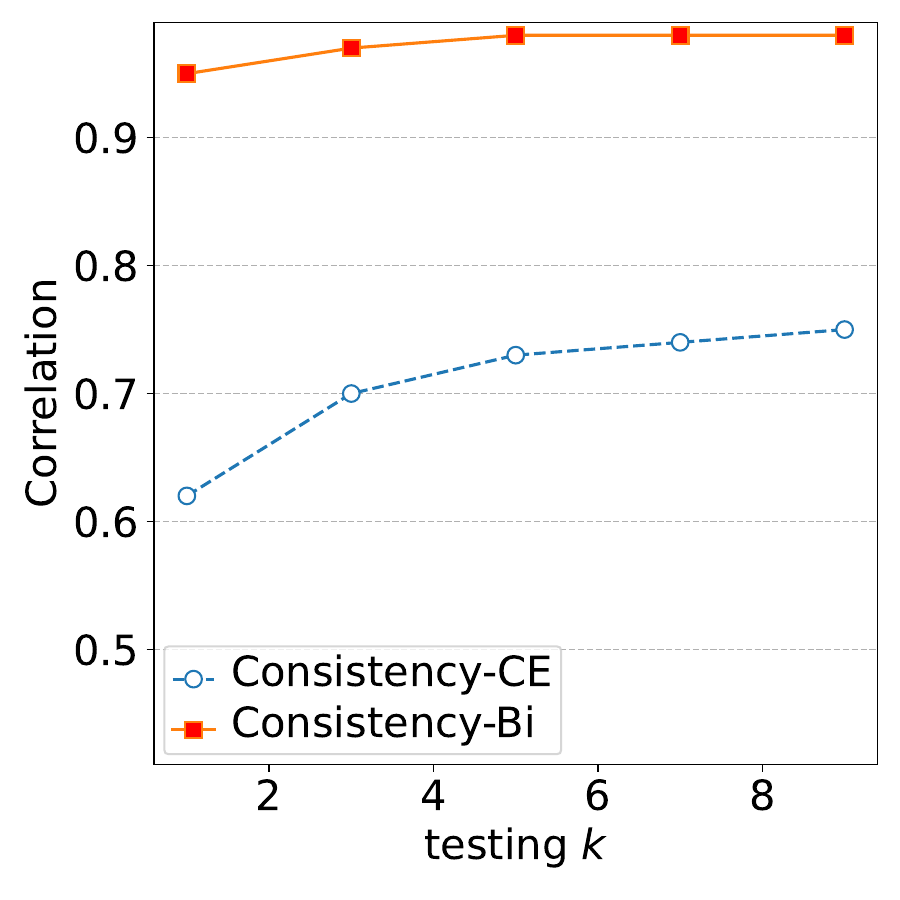}
\caption {Comparison of correlation between training loss (cross-entropy objective loss for Consistency-CE and Bi-objective loss for Consistency-Bi) and translation quality (BLEU score) on train subset of WMT15 De-En task for wait-$k$ policy.}
\label{analysis.mrt_loss_more_consistency}
\end{figure}
\paragraph{Correlation between training loss and translation quality}
We analyze the correlation between training loss and BLEU score, similar to the analysis described in \S \ref{analysis.correlation_between_BLEU_and_celoss}. 
The results shown in Figure~\ref{analysis.mrt_loss_more_consistency} demonstrates that, compared with Consistency-CE, proposed Consistency-Bi exhibits a strong correlation between training loss and translation quality, even when using a small $k$.

\begin{figure}[t]
\centering
\includegraphics[width=1.8in]{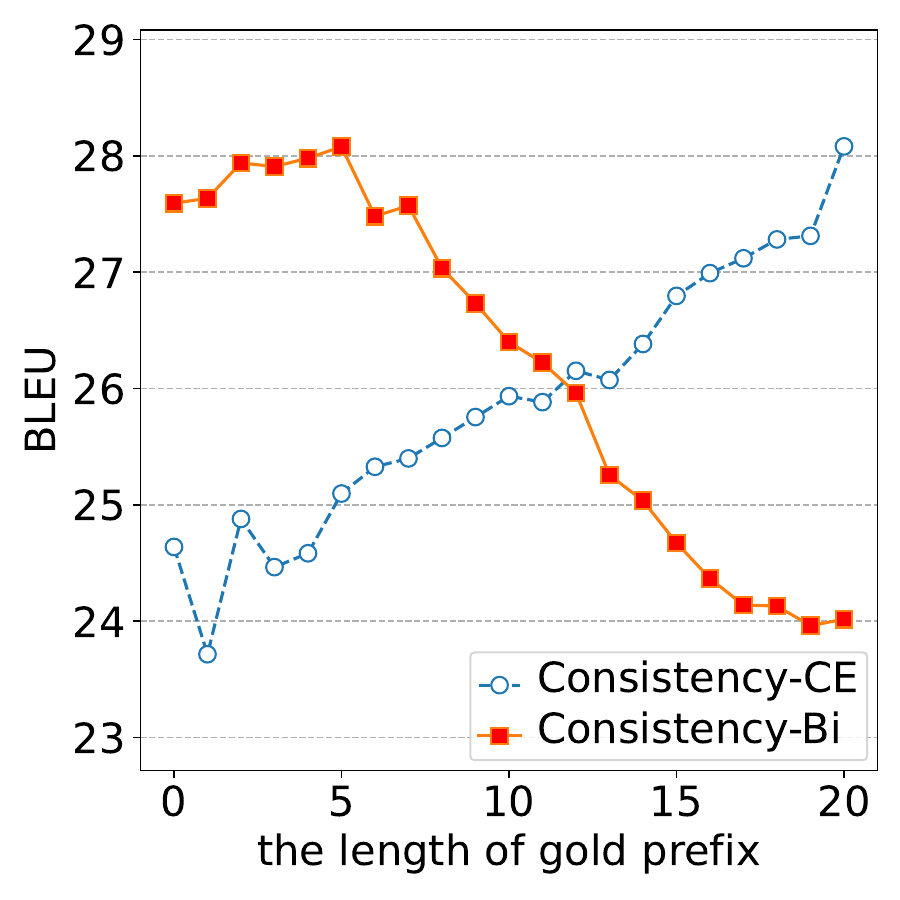}
\caption {BLEU comparison between original Consistency-CE model and ours proposed Consistency-Bi model for wait-$1$ testing under prefix-constrained decoding setting.}
\label{analysis.pseudo_origin_and_ours}
\end{figure}
\paragraph{Exposure Bias}
To assess whether our method successfully mitigates exposure bias discussed in \S \ref{sec:reason_of_problem}, we conduct wait-$1$ testing experiments using both original consistent wait-$1$ model (Consistency-CE) and proposed consistent wait-$1$ model (Consistency-Bi) under the prefix-constrained decoding setting \cite{wuebker2016models}. 
The detailed experimental settings are as described in \S \ref{sec:reason_of_problem}.
The results, presented in Figure \ref{analysis.pseudo_origin_and_ours}, reveal that as the number of gold prefix decreases, the performance of Consistency-Bi improves, while the performance of Consistency-CE deteriorates.
This suggests that proposed method effectively mitigates exposure bias, enhancing the model's performance when relying on prediction rather than on gold prefix.
In contrast, Consistency-CE exhibits serious exposure bias when overly dependent on gold prefix.


\section{Related Work}
\paragraph{SiMT Policy}
Read/write policies can be broadly categorized into two types: fixed and adaptive.
In the domain of fixed policies,  \citet{dalvi-etal-2018-incremental} introduced STATIC-RW, and \citet{ma-etal-2019-stacl} proposed the wait-$k$ policy, which consistently generates target tokens lagging behind the source by $k$ positions. Building upon this, \citet{elbayad2020efficient} enhanced the wait-$k$ policy by introducing the practice of sampling different values of $k$ during training. Additionally, \citet{han-etal-2020-end} incorporated meta-learning into the wait-$k$ policy, and \citet{zhang2021future} proposed future-guided training for the wait-$k$ policy. 
Shifting to adaptive policies, \citet{zheng_simultaneous_2020} developed an adaptive wait-$k$ policy through a heuristic ensemble of multiple wait-$k$ models. \citet{Zheng2019b} trained an agent with a gold read/write sequence. MILk \cite{arivazhagan2019monotonic} predicts a Bernoulli variable to determine READ/WRITE actions, which is further implemented into the transformer architecture MMA \cite{ma_monotonic_2020}. Additionally, \citet{zhang-zhang-2020-dynamic} and \citet{zhang-etal-2020-learning-adaptive} proposed adaptive segmentation policies, while alignment-based chunking policies were introduced by \citet{bahar-etal-2020-start} and \citet{wilken-etal-2020-neural}. \citet{miao2021generative} proposed a generative framework for generating READ/WRITE decisions. Furthermore, \citet{zhang_information-transport-based_2022} developed a READ/WRITE policy by modeling the translation process as information transport. \citet{zhang2022wait} developed a READ/WRITE policy based on the comparison between the total information of the previous target and the received source. 

\paragraph{SiMT Training Approach}
In the realm of SiMT training approaches, the training of early works in simultaneous translation \cite{bangalore-etal-2012-real,gu-etal-2017-learning} is based on full-sentence context, despite their testing scenarios involving partial context.
Addressing this disparity between full training context and partial testing context, \citet{ma2018stacl} introduced partial context training, also known as Prefix-to-Prefix Training, which is a context consistency training.
Building on this, \citet{elbayad2020efficient} proposed multi-path training, involving the sampling of different values of $k$ during training, a concept subsequently embraced by \citet{zhang2021universal,zhang2022wait,guo2022turning,zhang_information-transport-based_2022,wang2023better,guo2023glancing}.
Expanding on the concept of multi-path training, \citet{zhang_information-transport-based_2022} introduced curriculum-based training, following an easy-to-hard schedule. Additionally, \citet{guo2022turning} proposed disturbed-path training, while \citet{guo2023glancing} proposed the glancing future training, which involves introducing future source information.



\section{Conclusion}
In this paper, we pay attention to a counterintuitive phenomenon on the context usage between training and testing in SiMT. 
Subsequently, we conduct the comprehensive analysis and make the noteworthy discovery that
this phenomenon primarily stems from the weak correlation between translation quality and training loss as well as exposure bias between training and testing.
Based on our findings, we accordingly propose context consistency training method that incorporates both translation quality and latency as bi-objective and alleviates the exposure bias issue during the training stage of SiMT. 
Experiments on three language pairs and several SiMT Systems demonstrate the effectiveness of our proposed approach, making the context consistent model successful for the first time.

\section*{Limitations}
Our context consistency training approach necessitates a search for an appropriate hyperparameter, denoted as $\gamma$, to strike a balance between translation quality and latency. Further research is required to establish an efficient method for this purpose.

\bibliography{CCT4SiMT}

\begin{thebibliography}{43}
\expandafter\ifx\csname natexlab\endcsname\relax\def\natexlab#1{#1}\fi

\bibitem[{Arivazhagan et~al.(2019)Arivazhagan, Cherry, Macherey, Chiu, Yavuz, Pang, Li, and Raffel}]{arivazhagan2019monotonic}
Naveen Arivazhagan, Colin Cherry, Wolfgang Macherey, Chung-Cheng Chiu, Semih Yavuz, Ruoming Pang, Wei Li, and Colin Raffel. 2019.
\newblock Monotonic infinite lookback attention for simultaneous machine translation.
\newblock \emph{arXiv preprint arXiv:1906.05218}.

\bibitem[{Bahar et~al.(2020)Bahar, Wilken, Alkhouli, Guta, Golik, Matusov, and Herold}]{bahar-etal-2020-start}
Parnia Bahar, Patrick Wilken, Tamer Alkhouli, Andreas Guta, Pavel Golik, Evgeny Matusov, and Christian Herold. 2020.
\newblock \href {https://doi.org/10.18653/v1/2020.iwslt-1.3} {Start-before-end and end-to-end: Neural speech translation by {A}pp{T}ek and {RWTH} {A}achen {U}niversity}.
\newblock In \emph{Proceedings of the 17th International Conference on Spoken Language Translation}, pages 44--54, Online. Association for Computational Linguistics.

\bibitem[{Bangalore et~al.(2012)Bangalore, Rangarajan~Sridhar, Kolan, Golipour, and Jimenez}]{bangalore-etal-2012-real}
Srinivas Bangalore, Vivek~Kumar Rangarajan~Sridhar, Prakash Kolan, Ladan Golipour, and Aura Jimenez. 2012.
\newblock \href {https://aclanthology.org/N12-1048} {Real-time incremental speech-to-speech translation of dialogs}.
\newblock In \emph{Proceedings of the 2012 Conference of the North {A}merican Chapter of the Association for Computational Linguistics: Human Language Technologies}, pages 437--445, Montr{\'e}al, Canada. Association for Computational Linguistics.

\bibitem[{Bengio et~al.(2015)Bengio, Vinyals, Jaitly, and Shazeer}]{bengio2015scheduled}
Samy Bengio, Oriol Vinyals, Navdeep Jaitly, and Noam Shazeer. 2015.
\newblock Scheduled sampling for sequence prediction with recurrent neural networks.
\newblock \emph{Advances in neural information processing systems}, 28.

\bibitem[{Callison-Burch et~al.(2009)Callison-Burch, Koehn, Monz, and Schroeder}]{callison-burch-etal-2009-findings}
Chris Callison-Burch, Philipp Koehn, Christof Monz, and Josh Schroeder. 2009.
\newblock \href {https://aclanthology.org/W09-0401} {Findings of the 2009 {W}orkshop on {S}tatistical {M}achine {T}ranslation}.
\newblock In \emph{Proceedings of the Fourth Workshop on Statistical Machine Translation}, pages 1--28, Athens, Greece. Association for Computational Linguistics.

\bibitem[{Cettolo et~al.(2014)Cettolo, Niehues, St{\"u}ker, Bentivogli, and Federico}]{Cettolo14iwslt}
Mauro Cettolo, Jan Niehues, Sebastian St{\"u}ker, Luisa Bentivogli, and Marcello Federico. 2014.
\newblock \href {http://isl.anthropomatik.kit.edu/cmu-kit/downloads/Report_on_the_11th_IWSLT_Evaluation_Campaign_IWSLT_2014.pdf} {Report on the 11th {IWSLT} evaluation campaign}.
\newblock In \emph{iwslt}.

\bibitem[{Cho and Esipova(2016)}]{Cho2016}
Kyunghyun Cho and Masha Esipova. 2016.
\newblock \href {http://arxiv.org/abs/1606.02012} {{Can neural machine translation do simultaneous translation?}}

\bibitem[{Dalvi et~al.(2018)Dalvi, Durrani, Sajjad, and Vogel}]{dalvi-etal-2018-incremental}
Fahim Dalvi, Nadir Durrani, Hassan Sajjad, and Stephan Vogel. 2018.
\newblock \href {https://doi.org/10.18653/v1/N18-2079} {Incremental decoding and training methods for simultaneous translation in neural machine translation}.
\newblock In \emph{Proceedings of the 2018 Conference of the North {A}merican Chapter of the Association for Computational Linguistics: Human Language Technologies, Volume 2 (Short Papers)}, pages 493--499, New Orleans, Louisiana. Association for Computational Linguistics.

\bibitem[{Edunov et~al.(2017)Edunov, Ott, Auli, Grangier, and Ranzato}]{edunov2017classical}
Sergey Edunov, Myle Ott, Michael Auli, David Grangier, and Marc'Aurelio Ranzato. 2017.
\newblock Classical structured prediction losses for sequence to sequence learning.
\newblock \emph{arXiv preprint arXiv:1711.04956}.

\bibitem[{Elbayad et~al.(2020)Elbayad, Besacier, and Verbeek}]{elbayad2020efficient}
Maha Elbayad, Laurent Besacier, and Jakob Verbeek. 2020.
\newblock Efficient wait-k models for simultaneous machine translation.
\newblock \emph{arXiv preprint arXiv:2005.08595}.

\bibitem[{Gu et~al.(2017)Gu, Neubig, Cho, and Li}]{gu-etal-2017-learning}
Jiatao Gu, Graham Neubig, Kyunghyun Cho, and Victor~O.K. Li. 2017.
\newblock \href {https://aclanthology.org/E17-1099} {Learning to translate in real-time with neural machine translation}.
\newblock In \emph{Proceedings of the 15th Conference of the {E}uropean Chapter of the Association for Computational Linguistics: Volume 1, Long Papers}, pages 1053--1062, Valencia, Spain. Association for Computational Linguistics.

\bibitem[{Guo et~al.(2022)Guo, Zhang, and Feng}]{guo2022turning}
Shoutao Guo, Shaolei Zhang, and Yang Feng. 2022.
\newblock Turning fixed to adaptive: Integrating post-evaluation into simultaneous machine translation.
\newblock \emph{arXiv preprint arXiv:2210.11900}.

\bibitem[{Guo et~al.(2023)Guo, Zhang, and Feng}]{guo2023glancing}
Shoutao Guo, Shaolei Zhang, and Yang Feng. 2023.
\newblock Glancing future for simultaneous machine translation.
\newblock \emph{arXiv preprint arXiv:2309.06179}.

\bibitem[{Han et~al.(2020)Han, Zaidi, Indurthi, Lakumarapu, Lee, and Kim}]{han-etal-2020-end}
Hou~Jeung Han, Mohd~Abbas Zaidi, Sathish~Reddy Indurthi, Nikhil~Kumar Lakumarapu, Beomseok Lee, and Sangha Kim. 2020.
\newblock \href {https://doi.org/10.18653/v1/2020.iwslt-1.5} {End-to-end simultaneous translation system for {IWSLT}2020 using modality agnostic meta-learning}.
\newblock In \emph{Proceedings of the 17th International Conference on Spoken Language Translation}, pages 62--68, Online. Association for Computational Linguistics.

\bibitem[{Holtzman et~al.(2019)Holtzman, Buys, Du, Forbes, and Choi}]{holtzman2019curious}
Ari Holtzman, Jan Buys, Li~Du, Maxwell Forbes, and Yejin Choi. 2019.
\newblock The curious case of neural text degeneration.
\newblock \emph{arXiv preprint arXiv:1904.09751}.

\bibitem[{Kudo and Richardson(2018)}]{kudo_sentencepiece_2018}
Taku Kudo and John Richardson. 2018.
\newblock \href {https://doi.org/10.18653/v1/D18-2012} {{SentencePiece}: {A} simple and language independent subword tokenizer and detokenizer for {Neural} {Text} {Processing}}.
\newblock In \emph{Proceedings of the 2018 {Conference} on {Empirical} {Methods} in {Natural} {Language} {Processing}: {System} {Demonstrations}}, pages 66--71, Brussels, Belgium. Association for Computational Linguistics.

\bibitem[{Luong and Manning(2015)}]{Luong15iwslt}
Minh-Thang Luong and Christopher~D. Manning. 2015.
\newblock \href {https://nlp.stanford.edu/pubs/luong-manning-iwslt15.pdf} {Stanford neural machine translation systems for spoken language domains}.

\bibitem[{Ma et~al.(2019)Ma, Huang, Xiong, Zheng, Liu, Zheng, Zhang, He, Liu, Li, Wu, and Wang}]{ma-etal-2019-stacl}
Mingbo Ma, Liang Huang, Hao Xiong, Renjie Zheng, Kaibo Liu, Baigong Zheng, Chuanqiang Zhang, Zhongjun He, Hairong Liu, Xing Li, Hua Wu, and Haifeng Wang. 2019.
\newblock \href {https://doi.org/10.18653/v1/P19-1289} {{STACL}: Simultaneous translation with implicit anticipation and controllable latency using prefix-to-prefix framework}.
\newblock In \emph{Proceedings of the 57th Annual Meeting of the Association for Computational Linguistics}, pages 3025--3036, Florence, Italy. Association for Computational Linguistics.

\bibitem[{Ma et~al.(2018)Ma, Huang, Xiong, Zheng, Liu, Zheng, Zhang, He, Liu, Li et~al.}]{ma2018stacl}
Mingbo Ma, Liang Huang, Hao Xiong, Renjie Zheng, Kaibo Liu, Baigong Zheng, Chuanqiang Zhang, Zhongjun He, Hairong Liu, Xing Li, et~al. 2018.
\newblock Stacl: Simultaneous translation with implicit anticipation and controllable latency using prefix-to-prefix framework.
\newblock \emph{arXiv preprint arXiv:1810.08398}.

\bibitem[{Ma et~al.(2020)Ma, Pino, Cross, Puzon, and Gu}]{ma_monotonic_2020}
Xutai Ma, Juan~Miguel Pino, James Cross, Liezl Puzon, and Jiatao Gu. 2020.
\newblock \href {https://openreview.net/forum?id=Hyg96gBKPS} {Monotonic {Multihead} {Attention}}.

\bibitem[{Miao et~al.(2021)Miao, Blunsom, and Specia}]{miao2021generative}
Yishu Miao, Phil Blunsom, and Lucia Specia. 2021.
\newblock A generative framework for simultaneous machine translation.
\newblock In \emph{Proceedings of the 2021 Conference on Empirical Methods in Natural Language Processing}, pages 6697--6706.

\bibitem[{Ott et~al.(2019)Ott, Edunov, Baevski, Fan, Gross, Ng, Grangier, and Auli}]{ott-etal-2019-fairseq}
Myle Ott, Sergey Edunov, Alexei Baevski, Angela Fan, Sam Gross, Nathan Ng, David Grangier, and Michael Auli. 2019.
\newblock \href {https://doi.org/10.18653/v1/N19-4009} {fairseq: A fast, extensible toolkit for sequence modeling}.
\newblock In \emph{Proceedings of the 2019 Conference of the North {A}merican Chapter of the Association for Computational Linguistics (Demonstrations)}, pages 48--53, Minneapolis, Minnesota. Association for Computational Linguistics.

\bibitem[{Papineni et~al.(2002{\natexlab{a}})Papineni, Roukos, Ward, and Zhu}]{papineni2002bleu}
Kishore Papineni, Salim Roukos, Todd Ward, and Wei-Jing Zhu. 2002{\natexlab{a}}.
\newblock Bleu: a method for automatic evaluation of machine translation.
\newblock In \emph{Proceedings of the 40th annual meeting of the Association for Computational Linguistics}, pages 311--318.

\bibitem[{Papineni et~al.(2002{\natexlab{b}})Papineni, Roukos, Ward, and Zhu}]{papineni-etal-2002-bleu}
Kishore Papineni, Salim Roukos, Todd Ward, and Wei-Jing Zhu. 2002{\natexlab{b}}.
\newblock \href {https://doi.org/10.3115/1073083.1073135} {{B}leu: a method for automatic evaluation of machine translation}.
\newblock In \emph{Proceedings of the 40th Annual Meeting of the Association for Computational Linguistics}, pages 311--318, Philadelphia, Pennsylvania, USA. Association for Computational Linguistics.

\bibitem[{Ranzato et~al.(2015)Ranzato, Chopra, Auli, and Zaremba}]{ranzato2015sequence}
Marc'Aurelio Ranzato, Sumit Chopra, Michael Auli, and Wojciech Zaremba. 2015.
\newblock Sequence level training with recurrent neural networks.
\newblock \emph{arXiv preprint arXiv:1511.06732}.

\bibitem[{Sennrich et~al.(2016)Sennrich, Haddow, and Birch}]{Sennrich16acl}
Rico Sennrich, Barry Haddow, and Alexandra Birch. 2016.
\newblock \href {https://www.aclweb.org/anthology/P16-1162/} {Neural machine translation of rare words with subword units}.
\newblock In \emph{acl}.

\bibitem[{Shen et~al.(2016)Shen, Cheng, He, He, Wu, Sun, and Liu}]{shen2016minimum}
Shiqi Shen, Yong Cheng, Zhongjun He, Wei He, Hua Wu, Maosong Sun, and Yang Liu. 2016.
\newblock Minimum risk training for neural machine translation.
\newblock In \emph{Proceedings of the 54th Annual Meeting of the Association for Computational Linguistics (Volume 1: Long Papers)}, pages 1683--1692.

\bibitem[{Vaswani et~al.(2017)Vaswani, Shazeer, Parmar, Uszkoreit, Jones, Gomez, Kaiser, and Polosukhin}]{vaswani2017attention}
Ashish Vaswani, Noam Shazeer, Niki Parmar, Jakob Uszkoreit, Llion Jones, Aidan~N Gomez, {\L}ukasz Kaiser, and Illia Polosukhin. 2017.
\newblock Attention is all you need.
\newblock \emph{Advances in neural information processing systems}, 30.

\bibitem[{Wang et~al.(2023)Wang, Wu, Fan, Luo, Xiao, and Huang}]{wang2023better}
Shushu Wang, Jing Wu, Kai Fan, Wei Luo, Jun Xiao, and Zhongqiang Huang. 2023.
\newblock Better simultaneous translation with monotonic knowledge distillation.
\newblock In \emph{Proceedings of the 61st Annual Meeting of the Association for Computational Linguistics (Volume 1: Long Papers)}, pages 2334--2349.

\bibitem[{Wieting et~al.(2019)Wieting, Berg-Kirkpatrick, Gimpel, and Neubig}]{wieting2019beyond}
John Wieting, Taylor Berg-Kirkpatrick, Kevin Gimpel, and Graham Neubig. 2019.
\newblock Beyond bleu: Training neural machine translation with semantic similarity.
\newblock In \emph{Proceedings of the 57th Annual Meeting of the Association for Computational Linguistics}, pages 4344--4355.

\bibitem[{Wilken et~al.(2020)Wilken, Alkhouli, Matusov, and Golik}]{wilken-etal-2020-neural}
Patrick Wilken, Tamer Alkhouli, Evgeny Matusov, and Pavel Golik. 2020.
\newblock \href {https://doi.org/10.18653/v1/2020.iwslt-1.29} {Neural simultaneous speech translation using alignment-based chunking}.
\newblock In \emph{Proceedings of the 17th International Conference on Spoken Language Translation}, pages 237--246, Online. Association for Computational Linguistics.

\bibitem[{Wuebker et~al.(2016)Wuebker, Green, DeNero, Hasan, and Luong}]{wuebker2016models}
Joern Wuebker, Spence Green, John DeNero, Sa{\v{s}}a Hasan, and Minh-Thang Luong. 2016.
\newblock Models and inference for prefix-constrained machine translation.
\newblock In \emph{Proceedings of the 54th Annual Meeting of the Association for Computational Linguistics (Volume 1: Long Papers)}, pages 66--75.

\bibitem[{Zhang and Zhang(2020)}]{zhang-zhang-2020-dynamic}
Ruiqing Zhang and Chuanqiang Zhang. 2020.
\newblock \href {https://doi.org/10.18653/v1/2020.autosimtrans-1.1} {Dynamic sentence boundary detection for simultaneous translation}.
\newblock In \emph{Proceedings of the First Workshop on Automatic Simultaneous Translation}, pages 1--9, Seattle, Washington. Association for Computational Linguistics.

\bibitem[{Zhang et~al.(2020)Zhang, Zhang, He, Wu, and Wang}]{zhang-etal-2020-learning-adaptive}
Ruiqing Zhang, Chuanqiang Zhang, Zhongjun He, Hua Wu, and Haifeng Wang. 2020.
\newblock \href {https://doi.org/10.18653/v1/2020.emnlp-main.178} {Learning adaptive segmentation policy for simultaneous translation}.
\newblock In \emph{Proceedings of the 2020 Conference on Empirical Methods in Natural Language Processing (EMNLP)}, pages 2280--2289, Online. Association for Computational Linguistics.

\bibitem[{Zhang and Feng(2021)}]{zhang2021universal}
Shaolei Zhang and Yang Feng. 2021.
\newblock Universal simultaneous machine translation with mixture-of-experts wait-k policy.
\newblock \emph{arXiv preprint arXiv:2109.05238}.

\bibitem[{Zhang and Feng(2022{\natexlab{a}})}]{zhang_information-transport-based_2022}
Shaolei Zhang and Yang Feng. 2022{\natexlab{a}}.
\newblock \href {http://arxiv.org/abs/2210.12357} {Information-{Transport}-based {Policy} for {Simultaneous} {Translation}}.
\newblock ArXiv:2210.12357 [cs, eess].

\bibitem[{Zhang and Feng(2022{\natexlab{b}})}]{zhang-feng-2022-modeling}
Shaolei Zhang and Yang Feng. 2022{\natexlab{b}}.
\newblock \href {https://doi.org/10.18653/v1/2022.acl-long.176} {Modeling dual read/write paths for simultaneous machine translation}.
\newblock In \emph{Proceedings of the 60th Annual Meeting of the Association for Computational Linguistics (Volume 1: Long Papers)}, pages 2461--2477, Dublin, Ireland. Association for Computational Linguistics.

\bibitem[{Zhang and Feng(2022{\natexlab{c}})}]{zhang-feng-2022-reducing}
Shaolei Zhang and Yang Feng. 2022{\natexlab{c}}.
\newblock \href {https://doi.org/10.18653/v1/2022.acl-long.467} {Reducing position bias in simultaneous machine translation with length-aware framework}.
\newblock In \emph{Proceedings of the 60th Annual Meeting of the Association for Computational Linguistics (Volume 1: Long Papers)}, pages 6775--6788, Dublin, Ireland. Association for Computational Linguistics.

\bibitem[{Zhang et~al.(2021)Zhang, Feng, and Li}]{zhang2021future}
Shaolei Zhang, Yang Feng, and Liangyou Li. 2021.
\newblock Future-guided incremental transformer for simultaneous translation.
\newblock In \emph{Proceedings of the AAAI Conference on Artificial Intelligence}, volume~35, pages 14428--14436.

\bibitem[{Zhang et~al.(2022)Zhang, Guo, and Feng}]{zhang2022wait}
Shaolei Zhang, Shoutao Guo, and Yang Feng. 2022.
\newblock Wait-info policy: Balancing source and target at information level for simultaneous machine translation.
\newblock \emph{arXiv preprint arXiv:2210.11220}.

\bibitem[{Zhang et~al.(2019)Zhang, Feng, Meng, You, and Liu}]{zhang-etal-2019-bridging}
Wen Zhang, Yang Feng, Fandong Meng, Di~You, and Qun Liu. 2019.
\newblock \href {https://doi.org/10.18653/v1/P19-1426} {Bridging the gap between training and inference for neural machine translation}.
\newblock In \emph{Proceedings of the 57th Annual Meeting of the Association for Computational Linguistics}, pages 4334--4343, Florence, Italy. Association for Computational Linguistics.

\bibitem[{Zheng et~al.(2020)Zheng, Liu, Zheng, Ma, Liu, and Huang}]{zheng_simultaneous_2020}
Baigong Zheng, Kaibo Liu, Renjie Zheng, Mingbo Ma, Hairong Liu, and Liang Huang. 2020.
\newblock \href {https://doi.org/10.18653/v1/2020.acl-main.254} {Simultaneous {Translation} {Policies}: {From} {Fixed} to {Adaptive}}.
\newblock In \emph{Proceedings of the 58th {Annual} {Meeting} of the {Association} for {Computational} {Linguistics}}, pages 2847--2853, Online. Association for Computational Linguistics.

\bibitem[{Zheng et~al.(2019)Zheng, Zheng, Ma, and Huang}]{Zheng2019b}
Baigong Zheng, Renjie Zheng, Mingbo Ma, and Liang Huang. 2019.
\newblock \href {https://doi.org/10.18653/v1/D19-1137} {Simpler and faster learning of adaptive policies for simultaneous translation}.
\newblock In \emph{Proceedings of the 2019 Conference on Empirical Methods in Natural Language Processing and the 9th International Joint Conference on Natural Language Processing (EMNLP-IJCNLP)}, pages 1349--1354, Hong Kong, China. Association for Computational Linguistics.

\end{thebibliography}
\bibliographystyle{acl_natbib}

\end{document}